\documentclass[11pt]{article}
\usepackage[final]{acl}
\usepackage{times}
\usepackage{latexsym}
\usepackage[T1]{fontenc}
\usepackage[utf8]{inputenc}
\usepackage{microtype}
\usepackage{inconsolata}
\usepackage{graphicx}
\usepackage{amsfonts}
\usepackage{placeins}
\usepackage{amssymb}
\usepackage{amsmath}
\DeclareMathOperator{\softmax}{softmax}
\newcommand{\N}{\mathcal{N}}
\newcommand{\NN}{\mathrm{NN}}
\newcommand{\Cat}{\mathrm{Cat}}

\newcommand{\E}{\mathbb{E}}
\newcommand{\elbo}{\mathcal{L}}
\usepackage{float}
\usepackage{subcaption}
\usepackage{array}
\usepackage{booktabs}
\usepackage{algorithmicx}
\usepackage{algcompatible}
\usepackage{algorithm}
\usepackage{algpseudocode}
\usepackage{multirow}
\usepackage{soul}
\usepackage{dblfloatfix} 
\usepackage{balance}     
\usepackage{multicol}
\usepackage{caption}

\usepackage{cuted}
\usepackage{capt-of}

\title{Modeling Document and Section Hierarchies in Embedding Space}

\author{\textbf{Kriste Krstovski}, \textbf{Sharyn O'Halloran}, \textbf{Kshitij Chandna}\\ 
Columbia University\\
\small{
\href{kriste.krstovski@columbia.edu}{kriste.krstovski@columbia.edu}
}
}

\begin{document}
\maketitle
\begin{abstract}
Embedded topic models integrate word embeddings into the topic generation process, producing coherent topics and robust performance on large vocabularies. However, each document is represented by a single topic distribution, ignoring its internal structure.  Many real world collections, including legal documents, company and government reports, and online discussion forums, exhibit hierarchical organization in which documents consist of multiple sections that often address distinct subtopics. We introduce the Hierarchical Embedded Topic Model (hETM), a structured topic model that couples the topic distribution of a document and those of its sections through a hierarchical prior in the embedding space. We derive an amortized variational inference procedure for hETM and evaluate it on structured collections from each of these domains. Experimental results show that hETM achieves the lowest or near-lowest held-out perplexity among flat and hierarchical topic model baselines, along with comparable or higher topic quality. It also performs strongly on a downstream clustering task, achieving the best clustering performance on most of the collections.

\end{abstract}

\section{Introduction}
\label{sec:intro}
Bag of words (BOW) models discard grammar and rely only on word co-occurrence statistics. Topic models, from latent Dirichlet allocation (LDA) \citep{blei2003lda} to recent neural variants \citep{dieng2020etm}, all adopt this representation. The way a corpus is split into documents therefore matters: long documents, such as books or statutes, yield diffuse statistics, while short ones, such as forum posts, yield sparsity. Yet many real world collections are explicitly hierarchical: a document is a structured sequence of sections, and meaning lives at both levels.

This structure is clearest in the \emph{legal} domain. Statutes and bills are divided into sections, and contracts into clauses: a single provision covers one topic, but its meaning depends on the bill it belongs to, since large bills often mix trade or tax provisions with unrelated ones. In \emph{finance}, company disclosures are organized into predefined sections, such as the ``Risk Factors'' or ``Management's Discussion and Analysis'' sections of a 10-K and the segments of an earnings call; analysts read these one at a time, for example to track a firm's liquidity risk, but still need a summary of the whole filing. In \emph{online forums}, a discussion thread is a sequence of short posts grouped into boards or newsgroups, as in the 20 Newsgroups collection \citep{lang1995newsweeder}, and a single post says little without the thread it belongs to. In each case, the section and the document are two levels of the same object, and a single topic distribution per document captures only one of them.

Treating such documents as flat bags of words forces a choice of unit at a cost: modeling the whole document blurs the distinct topics of its sections, while modeling each section on its own loses statistical strength. A hierarchy that ties each section's topics to its document recovers both: the section topics stay sharp, yet still combine into a document level theme. This matters most for long, varied documents with large, specialized vocabularies of legal citations, financial and scientific terms, and rare named entities --- exactly where flat LDA breaks down and embedding based topics help --- which calls for a model that captures both the document hierarchy and the vocabulary’s embedding structure.

Two largely separate lines of work address these limitations. The first incorporates word embeddings: the embedded topic model (ETM) of \citet{dieng2020etm} represents each topic as a point in a word embedding space and yields coherent topics on heavy-tailed vocabularies, and recent neural topic models build hierarchies of topics within that space \citep{duan2021sawetm,xu2022hyperminer,wu2024traco}. These models assume a single topic distribution per document. The second improves the structure of the prior: \citet{wallach2009rethinking} use asymmetric Dirichlet priors, and \citet{krstovski2016mlhlda} introduce mlhLDA, which couples per-section topic proportions through a document-level hierarchical Dirichlet prior but inherits LDA's bag of words likelihood and its degradation on large vocabularies. hETM focuses on the second kind of hierarchy. We model how a document’s topic proportions relate to those of its sections, rather than imposing a parent--child structure among the topics themselves. The topic-tree hierarchies cited above are complementary to this coupling and could be layered on top.

Combining embeddings with document-structure coupling is not straightforward. ETM places a logistic-normal prior on topic proportions and can be fit using the reparameterization trick \cite{kingma2014autoencoding}, while mlhLDA uses a nested Dirichlet prior that lets the document proportion serve as the base measure for the section. Swapping embedded topics under that prior forfeits ETM’s amortized inference. We propose the Hierarchical Embedded Topic Model (hETM): rather than importing mlhLDA’s Dirichlet hierarchy, we import its coupling, drawing each section proportion as a Gaussian perturbation of its document proportion in logit space. This preserves the shrinkage toward the document while keeping the model reparameterizable. Our contribution is twofold: (1) hETM, which couples section and document topic proportions through a hierarchical logistic-normal prior; and (2) an amortized variational inference procedure with a closed-form level-coupling term, in which the Gaussian precision plays the role of the Dirichlet concentration in mlhLDA.

We evaluate hETM on four collections across three domains: U.S.\ trade legislation and European Union (EU) legislation (legal), corporate annual reports (finance), and the 20 Newsgroups corpus (online forums). hETM attains lower perplexity than flat and hierarchical topic models, comparable or higher topic quality, and the highest clustering accuracy on most datasets. We review prior work, present hETM and its inference, and evaluate it intrinsically and extrinsically.
\section{Background and Related Work}
\label{sec:related}
Hierarchical prior structures for topic models have received attention in the past. For example, \citet{wallach2009rethinking} showed that an asymmetric Dirichlet prior, placed as a hyperprior over the document-topic distributions, provides better modeling of documents across the whole collection than the symmetric priors used in the original LDA. Other approaches build a tree over the topics themselves. The nested Chinese restaurant process \citep{blei2010ncrp} assigns each document a path down an unbounded topic tree, Pachinko allocation \citep{li2006pam} captures topic correlations through a directed acyclic graph of topics, and more recently \citet{kim2013variational} introduced tiLDA, a topic model of document collections with nested hierarchies. While these approaches model richer structure than LDA, they rely on Gibbs sampling or batch variational Bayes (VB) inference and, as such, require multiple iterations over the whole collection.

More closely related to our work, \citet{krstovski2016mlhlda} introduced mlhLDA, a topic model for documents with a predetermined section structure. Rather than a single document-topic distribution, mlhLDA assigns each section its own topic distribution and couples these to a document level distribution through a hierarchical Dirichlet prior, so that the topics of a section are shrunk toward the topics of the document that contains it. mlhLDA is trained with online VB \citep{hoffman2010online}, requires only a single pass over the collection, and extends to multilingual collections (mlhPLTM) by sharing a tuple-level distribution across languages. As with the approaches above, however, mlhLDA represents each topic as a separate distribution over the vocabulary and, like LDA, does not perform well on large vocabularies.

A separate line of work replaces the Dirichlet-multinomial of LDA with neural networks (NNs). For example, \citet{miao2016neural} and \citet{srivastava2017autoencoding} cast topic modeling as amortized variational inference, in which an inference network encodes a document's BOW and a decoder reconstructs it. More recently, \citet{dieng2020etm} introduced the embedded topic model (ETM), which represents each topic as a point in a word embedding space and draws words using the inner product between the topic embedding and the word embeddings. Because it relies on word embeddings, the ETM produces coherent topics even on large vocabularies, a regime where LDA degrades, and because it places a logistic-normal rather than a Dirichlet prior on the topic proportions, it can be trained with the reparameterization trick. A related embedding based model is Gaussian-LDA \citep{das2015gaussian}, which generates word embeddings rather than words; an earlier embedding-based topic model, TopicVec \citep{li2016topicvec}, also represents topics as points in the word embedding space and combines them with LDA-style Dirichlet topic proportions via variational expectation-maximization. Both models keep a single topic distribution per document, as does ETM.

Several recent neural topic models add hierarchy to this embedding based view; \citet{wu2024survey} give a recent survey of the area. \citet{isonuma2020tsntm} introduce a tree structured neural topic model that parameterizes an unbounded topic tree with doubly recurrent networks. \citet{duan2021sawetm} propose SawETM, which stacks topic embedding layers in a deep gamma belief network, and \citet{xu2022hyperminer} introduce HyperMiner, which places word and topic embeddings in hyperbolic space so that its tree-like geometry captures the hierarchy among topics. \citet{wu2024traco} propose TraCo, which learns a topic tree through a transport-plan dependency between parent and child topic distributions and separates topic hierarchy from topic diversity. A separate, non-generative line embeds document sections and clusters them to read off a hierarchy after the fact \citep{viegas2025contextual}; such approaches, however, do not define a likelihood or a principled coupling between levels.

Sub-document topic modeling, which assigns a topic distribution to each sub-document unit rather than to the document as a whole, is a separate line with its own history \citep{wang2009sentencetopic, balikas2016sentencetopic, li2017recurrent, jiang2019sentence}, with SubNTM \citep{nguyen2025subntm} as the most recent and the closest to our setting. SubNTM places the between-level coupling in the generative equation, through a document-independent noise vector, whereas hETM places it in the prior itself, drawing each section’s topic distribution around that of the document that contains it.

Unlike the neural models above, which build a hierarchy over the topics themselves, our model builds a hierarchy over the document sections. A topic hierarchy of the kind learned by \citet{xu2022hyperminer} or \citet{wu2024traco} is complementary and could in principle be layered on top. To our knowledge, ours is the first generative model in which the section and document levels share a single embedded topic space and are coupled in the prior.

\section{Modeling Topic Hierarchies in Embedding Space}
\label{sec:model}
hETM applies to documents that are split into sections, giving a parent--child relationship between a document and its parts. The split can be explicit, as in legislative bills, EU regulations, the items of company reports, or implicit, as when several short documents are grouped into a larger unit, such as forum posts on a common theme. We refer to these parts as sections throughout.

A document and its sections describe topics at two levels. The document as a whole has an overall theme, while each section focuses on a narrower set of topics: a trade bill is broadly about commerce, but one section defines terms, another sets tariff schedules, and a third lists penalties. A model that captures only the document level blurs these sections together, while one that treats each section on its own loses the shared subject that ties them. hETM models both levels jointly. Each section has its own distribution over topics, and these distributions are drawn around a single document distribution. Because the two levels are coupled rather than estimated separately, the section topics stay sharp while the document distribution emerges as a coherent summary of them.

We consider a collection of $D$ documents, where each document $d_i=(s_{i,1},\dots,s_{i,n_i})$ is an ordered list of $n_i$ sections, and each section is a BOW over a vocabulary of size $V$. The ordering indexes the sections but is not used by the model, which treats sections as conditionally independent given the document representation; incorporating section order would require a sequential prior and is beyond our scope. We denote $x_{i,j}$ as the word counts of section $s_{i,j}$ and $x_i=\sum_j x_{i,j}$ as the document counts, and model the collection with $K$ topics.

As in embedding based topic models, hETM represents each of the $K$ topics not as a separate distribution over the vocabulary but as a point $\alpha_k\in\mathbb{R}^L$ in a shared word embedding space, where $\rho\in\mathbb{R}^{L\times V}$ is the matrix of word embeddings. The topic-word distribution $\beta_k$ is the softmax of the inner products between the topic and word embeddings,

\begin{equation}
\beta_k=\softmax(\rho^\top\alpha_k), \label{eq:beta}
\end{equation}

which lets hETM produce coherent topics on large vocabularies.

\subsection{Generative Process}
\label{sec:genproc}

hETM generates a document together with its sections. Each document is summarized by a document-topic distribution, and each of its sections by its own section-topic distribution over the same $K$ topics. Rather than drawing these independently, we draw the section distributions as perturbations of the document distribution, tying the two levels together. To keep the model reparameterizable, we place this hierarchy in the logit space: the logit of a section is drawn from a Gaussian centered on the logit of its document, and the corresponding proportions are the softmax of that logit. For each document $d_i$, hETM assumes the following generative process:

\begin{itemize}\setlength{\itemsep}{1pt}
\item Choose a document logit $\delta_i^d\sim\N(0,\sigma_d^2 I)$ and set
$\theta_i^d=\softmax(\delta_i^d)$.
\item For each section $s_{i,j}$ in document $d_i$:
\begin{itemize}\setlength{\itemsep}{1pt}
\item Choose a section logit $\delta_{i,j}^s\sim\N(\delta_i^d,\sigma_s^2 I)$
and set $\theta_{i,j}^s=\softmax(\delta_{i,j}^s)$.
\item For each word $w$ in section $s_{i,j}$:
\begin{itemize}\setlength{\itemsep}{1pt}
\item Choose a topic $z\sim\Cat(\theta_{i,j}^s)$.
\item Choose a word $w\sim\Cat(\beta_z)$.
\end{itemize}
\end{itemize}
\end{itemize}

The document logit $\delta_i^d$ serves as the mean of the section logits $\delta_{i,j}^s$, so each section’s topics vary around the document’s. The section variance $\sigma_s^2$ controls the coupling: small values shrink sections toward the document, large values let them vary freely. 

As in embedding based topic models, the topic assignment $z$ can be marginalized out, so that the probability of a word depends only on the section logit,
\begin{equation}
p(w_{i,j,n}\mid\delta_{i,j}^s)=(\theta_{i,j}^s)^\top\beta_{\cdot,w_{i,j,n}},
\label{eq:wordlik}
\end{equation}
and the joint probability of a document and its logits is
\begin{equation}
\begin{split}
p(d_i,&\,\delta^d_i,\delta^s_{i,1:n_i})=\N(\delta_i^d;0,\sigma_d^2 I)\\
&\times\prod_{j}\N(\delta_{i,j}^s;\delta_i^d,\sigma_s^2 I)
\prod_{n}(\theta_{i,j}^s)^\top\beta_{\cdot,w_{i,j,n}}.
\end{split}
\label{eq:joint}
\end{equation}

\section{Amortized Variational Inference}
\label{sec:inference}
The posterior over the document and section logits is intractable, so we approximate it with amortized variational inference. We share two inference networks across the corpus: a document encoder that maps the normalized document counts $\bar x_i=x_i/N_i$ to a diagonal Gaussian over the document logit, and a section encoder that maps the normalized section counts $\bar x_{i,j}$ to a diagonal Gaussian over each section logit. The variational posterior mirrors the generative factorization,

\begin{equation}
\begin{split}
  q(\delta_i^d,\delta^s_{i,1:n_i})
  &= q_{\nu_d}(\delta_i^d\mid x_i)\\
  &\quad\times \prod_j q_{\nu_s}(\delta_{i,j}^s\mid x_{i,j}),
\end{split}
\end{equation}

\noindent with $(\mu_i^d,\ \log\!\big((\sigma_i^d)^2\big))=\NN_d(\bar x_i;\nu_d)$ and $(\mu_{i,j}^s,\log(\sigma_{i,j}^s)^2)=\NN_s(\bar x_{i,j};\nu_s)$. 
Because the document encoder reads the aggregate counts $\bar x_i$, the document posterior draws on the combined word statistics of all its sections.

We fit the encoder and model parameters by maximizing an evidence lower bound (ELBO). For each document, the bound separates into a reconstruction term and two KL terms, $\elbo_i=\mathrm{(R)}-\mathrm{(D)}-\mathrm{(S)}$, where the reconstruction term and the document KL term are
\begin{align}
  \mathrm{(R)}&=\sum_j\E_{q(\delta^s_{i,j})}\!\Big[\textstyle\sum_v x_{i,j,v}\log(\theta^s_{i,j})^\top\!\beta_{\cdot,v}\Big],\label{eq:recon}\\[2pt]
  \mathrm{(D)}&=\tfrac12\!\sum_k\!\Big[\tfrac{(\mu^d_{i,k})^2+(\sigma^d_{i,k})^2}{\sigma_d^2}-1-\log\tfrac{(\sigma^d_{i,k})^2}{\sigma_d^2}\Big].\label{eq:dockl}
\end{align}
The section term is the only nonstandard one: the prior on a section logit, $\N(\delta^d_i,\sigma_s^2 I)$, has a random mean $\delta^d_i$, so we take its expectation under $q(\delta^d_i)$. Using $\E[(\mu^s-\delta^d)^2]=(\mu^s-\mu^d)^2+(\sigma^d)^2$, the section KL term becomes
\begin{equation}
\begin{split}
  \mathrm{(S)}=\tfrac12\sum_{j}\sum_k\Big[&
   \tfrac{(\sigma^s_{i,j,k})^2+(\mu^s_{i,j,k}-\mu^d_{i,k})^2+(\sigma^d_{i,k})^2}{\sigma_s^2}\\[-1pt]
   &-1+\log\sigma_s^2-\log(\sigma^s_{i,j,k})^2\Big].
\end{split}
\label{eq:seckl}
\end{equation}
The cross term $(\mu^s-\mu^d)^2/\sigma_s^2$ pulls each section mean toward its document mean with strength $1/\sigma_s^2$, so a smaller $\sigma_s^2$ ties sections more tightly to the document; the term $(\sigma^d)^2/\sigma_s^2$ accounts for the document's own posterior uncertainty and makes the expression exact.

Both KL terms are available in closed form, while the reconstruction term is estimated with a single reparameterized sample per section, $\delta^s_{i,j}=\mu^s_{i,j}+\sigma^s_{i,j}\odot\epsilon$ with $\epsilon\sim\N(0,I)$. Since the word likelihood \eqref{eq:wordlik} depends only on the section logit, the document logit is never sampled and enters the bound only through the two KL terms, which removes one level of Monte Carlo noise. As in online VB, we optimize the bound on minibatches of documents, recomputing the topics $\beta$ from the embeddings via \eqref{eq:beta} at each step. For all experiments, we fix $\sigma_d^2 = \sigma_s^2 = 1$ rather than learning them, so that the model has no free coupling parameter to tune. Appendix~\ref{sec:sigma-ablation} ablates this choice on the U.S.\ Trade Statutes and EUR-Lex: $\sigma_s^2 = 1$ is near-optimal on EUR-Lex, and on the Trade Statutes it is a compromise between looser coupling (better perplexity) and tighter coupling (better topic quality). Appendix~\ref{sec:appendix-cost} discusses parameter counts and training time. Algorithm~\ref{alg:hetm} summarizes the full procedure.

\begin{algorithm}[t]
\caption{Topic modeling with hETM}
\label{alg:hetm}
\small
\begin{algorithmic}[1]
\State Initialize topics $\{\rho,\alpha_{1:K}\}$, encoders $\{\nu_d,\nu_s\}$, variances $\sigma_d^2,\sigma_s^2$
\For{iteration $t=1,2,\dots$}
  \State Compute $\beta_k=\mathrm{softmax}(\rho^\top\alpha_k)$ for each topic $k$
  \State Sample a minibatch $B$ of documents
  \For{each document $i\in B$}
    \State $(\mu_i^d,\log(\sigma_i^d)^2)\gets \NN_d(\bar x_i;\nu_d)$
    \For{each section $j=1,\dots,n_i$}
      \State $(\mu_{i,j}^s,\log(\sigma_{i,j}^s)^2)\gets \NN_s(\bar x_{i,j};\nu_s)$
      \State $\delta_{i,j}^s\gets \mu_{i,j}^s+\sigma_{i,j}^s\odot\epsilon,\ \epsilon\sim\N(0,I)$
      \State $\theta_{i,j}^s\gets \mathrm{softmax}(\delta_{i,j}^s)$
      \State $p(x_{i,j,v})\gets (\theta_{i,j}^s)^\top\beta_{\cdot,v}$ for each word $v$
    \EndFor
    \State $\elbo_i\gets \mathrm{(R)}-\mathrm{(D)}-\mathrm{(S)}$ \Comment{Eqs.~\eqref{eq:recon}--\eqref{eq:seckl}}
  \EndFor
  \State Update $\{\rho,\alpha_{1:K},\nu_d,\nu_s\}$ (and optionally $\sigma_d^2,\sigma_s^2$) by ascending $\sum_{i\in B}\elbo_i$
\EndFor
\end{algorithmic}
\end{algorithm}

\section{Experimental Setup}
\label{sec:experimental_setup}
We evaluate hETM both intrinsically and extrinsically. In both settings, we compare it against four baseline models: ETM \citep{dieng2020etm}, SawETM \citep{duan2021sawetm}, TSNTM \citep{isonuma2020tsntm}, and mlhLDA \citep{krstovski2016mlhlda}. The latter couples document- and section-level topic distributions through a hierarchical Dirichlet prior; because its topics are full multinomials rather than embeddings, the gap between them isolates the contribution of the embedding-based likelihood. We use ETM in two variants: the default version, which infers a single topic distribution over the whole document, denoted ETM(D); and a version that treats each section as a separate document and infers a topic distribution per section, denoted ETM(S). We use four corpora across three domains: legal, finance, and online forums.

\subsection{Corpora}
\label{sec:corpora}
Across the corpora, the hierarchical document structure is either explicit or implicit. For every corpus, we build the effective vocabulary with the Mallet package \citep{mccallum2002mallet} and apply the same preprocessing, standard for BOW latent variable models of text: we lowercase the text, remove punctuation, and discard any token containing a numeric character. We also remove stopwords and the 50 most frequent words, along with words that appear in fewer than three documents ($df(w)<3$) or are shorter than four characters ($|w|<4$). Below, we describe each corpus together with its section statistics and effective vocabulary size.

\begin{figure}[t]
  \centering
  \includegraphics[width=\columnwidth]{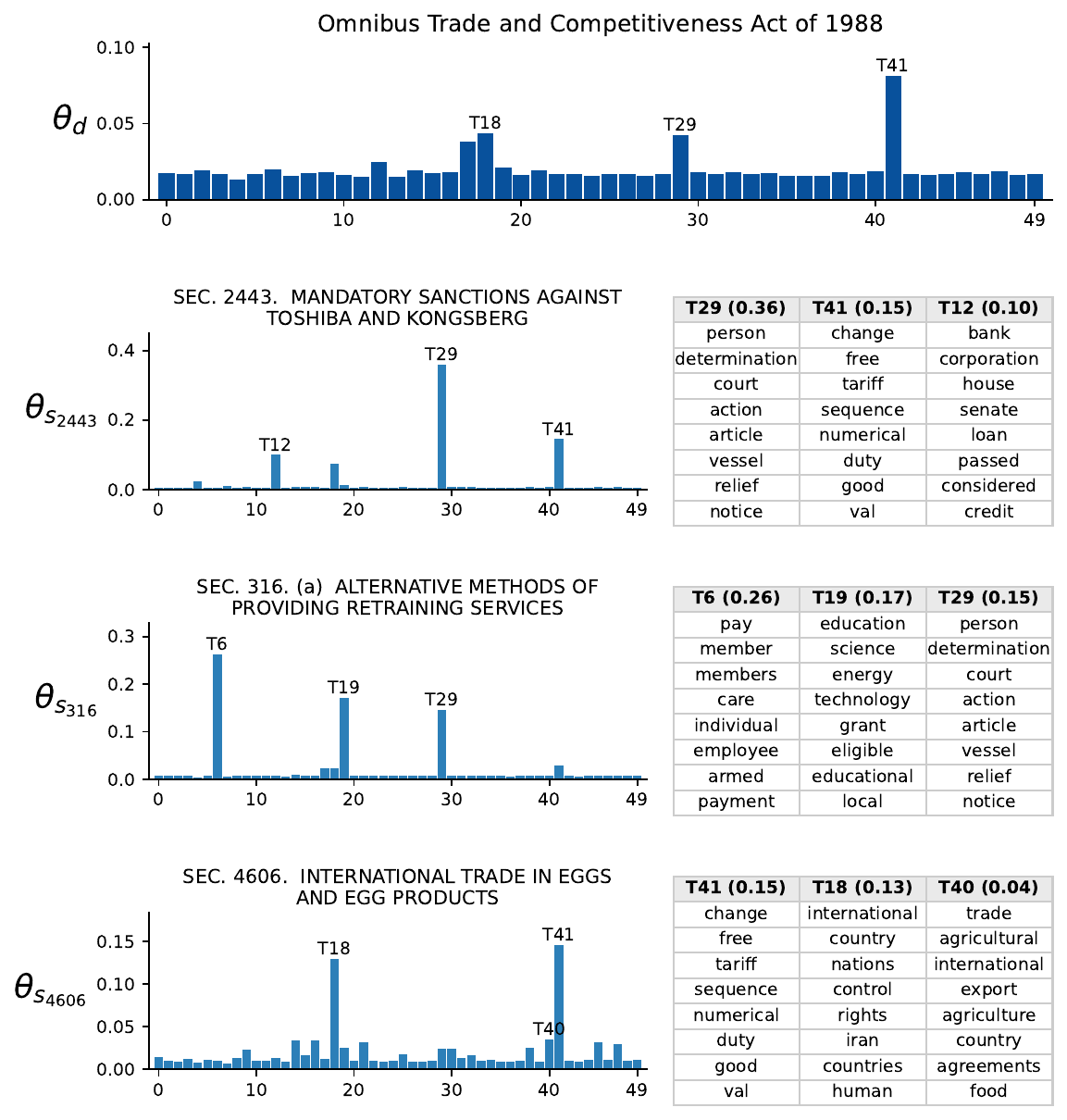}
\caption{hETM representation of the Omnibus Trade and Competitiveness Act of 1988 (50 topics): the document distribution $\theta_d$ and three example sections $\theta_s$ with the eight most probable words of their top topics.}
  \label{fig:usts_hetm_example_k50}
\end{figure}

\paragraph{U.S.\ trade legislation.}  
This is a corpus of U.S.\ trade legislation between 1951--2020. It consists of 576 bills comprising roughly 40k sections, with a heavy-tailed distribution (min 1, max 1{,}828, mean 69) that makes the section/document distinction sharp; 22\% of bills contain a single section. The effective vocabulary consists of 25,456 tokens. Figure~\ref{fig:usts_hetm_example_k50} shows an example hETM representation, with 50 topics, of the Omnibus Trade and Competitiveness Act of 1988. Shown at the top is the inferred topic distribution of the whole document, $\theta_d$, which in
hETM serves as the mean around which the section-topic distributions $\theta_{s}$ are drawn. Below it are three example sections (out of 475), each with its inferred distribution $\theta_{s}$ and the top eight words for its three most probable topics. The document distribution concentrates on a few topics, namely tariffs (T41), international relations and sanctions (T18), and legal enforcement (T29), and each section places its mass on the relevant subset, often blending several of them, so that $\theta_d$ is a smoothed aggregate of the sharper section distributions.

\paragraph{Edgar.} This is a financial corpus of annual reports (10-K filings) from all U.S.\ publicly traded companies, spanning 1993--2020 \citep{loukas-etal-2021-edgar}. We use the 2019 train split, which consists of 4,435 10-K reports. A 10-K is a standard form that companies must file annually, and it contains 15 numbered items. Since not all items contain text, we merge any item with fewer than 50 tokens into the preceding one. This yields 15,034 sections (1--4 per document, mean 3.39); 1.24\% of filings reduce to a single item after merging. Because 10-K reports are long and contain many company-specific tokens --- tickers, product and subsidiary names, personnel --- the default $df(w)<3$ cutoff used for the other corpora leaves a vocabulary dominated by rare, document-specific noise. We therefore raise the threshold to $df(w)\leq5$ on Edgar, yielding a vocabulary of 37{,}434 words that stays comparable in size to the other three corpora while keeping the words shared across companies.

\paragraph{EUR-Lex.} EUR-Lex is a corpus of English-language EU legislation, comprising regulations, directives, and decisions \citep{chalkidis-etal-2019-large}. We construct a subset of 10{,}000 documents, each split into its numbered sections, with an average document length of about 439 words. This yields 74{,}710 sections, with 4--46 sections per document (mean 7.47); every document has at least four sections, and the effective vocabulary is 14{,}163 words.

\paragraph{20 Newsgroups.} The 20 Newsgroups collection \citep{lang1995newsweeder} is a widely used corpus of Usenet posts spread across 20 newsgroups spanning domains such as computing, recreation, science, politics, and religion, some closely related and others unrelated. Our version contains 18{,}746 posts, ranging from 624 to 994 per newsgroup (mean 937.3), so no newsgroup collapses to a single section under the implicit hierarchy framing. We use this collection to illustrate an implicit hierarchy, treating each newsgroup as a document and its posts as sections, with an effective vocabulary of 18{,}627 words.

\subsection{Word Embeddings}
\label{sec:word_embeddings}
As with other embedding based topic models, hETM can either learn its own word embeddings jointly with the topics or use pretrained ones. Because word embeddings specialized for particular domains are now widely available, and our corpora span legal and financial text, we also evaluate hETM with several such pretrained models: LegalBERT \citep{chalkidis2020legalbert}, InCaseLawBERT \citep{paul2023indianlaw}, FinBERT \citep{araci2019finbert}, and ModernBERT \citep{warner2024modernbert}. Since these are contextual models that produce a different vector for each occurrence of a word, whereas hETM requires a single static embedding per vocabulary type, we convert each model into static embeddings by mean-pooling the contextual representation of each word over the corpus \citep{bommasani2020interpreting}. 

SawETM and TSNTM do not accept external pretrained embeddings and are run with their default learned embeddings.

\section{Intrinsic Evaluation}
\label{sec:intrinsic}

We first evaluate model fit using held-out perplexity. We randomly choose 10\% of the documents as a test set. For 20 Newsgroups, the held-out split is over posts within each newsgroup rather than at the parent-document level, so all 20 newsgroups appear in training. For each test document, we split each section into two halves. The first half of each section is used to infer the section-level topic distribution for section-based models (hETM, ETM(S), mlhLDA) and the document-level topic distribution for document-based models (ETM(D), SawETM, TSNTM). The inferred distribution is then used to compute perplexity over the words in the remaining half of that section.

\begin{figure*}[h]
  \centering
  \includegraphics[width=\textwidth]{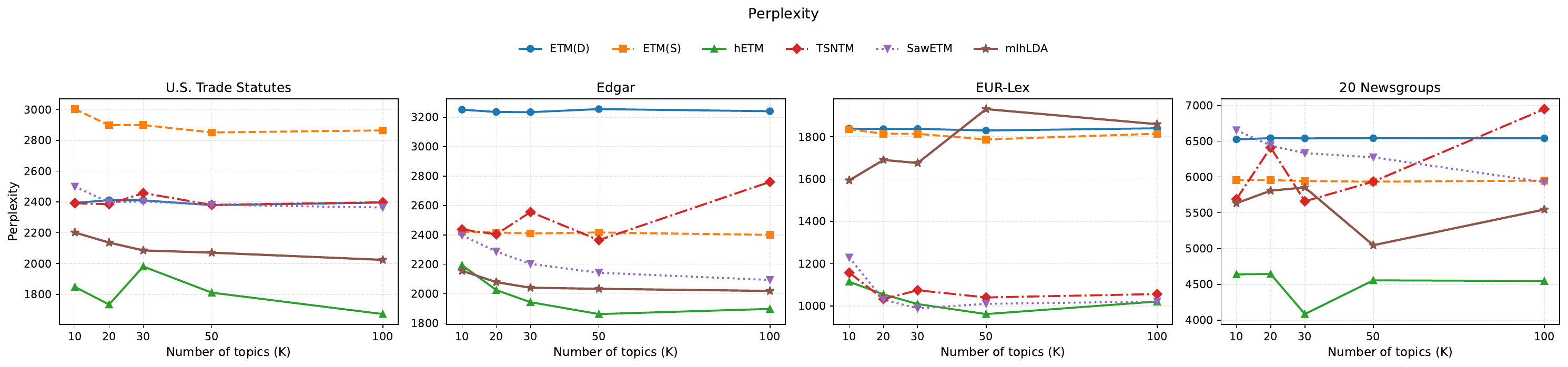}
  \caption{Held-out perplexity (section completion) as a function of
    the number of topics $K$ for ETM(D), ETM(S), hETM, SawETM, TSNTM, and mlhLDA on the four document corpora.}
  \label{fig:perplexity}
\end{figure*}

Figure~\ref{fig:perplexity} reports perplexity across the four collections. For each model and corpus we ran experiments with both self-trained and pretrained embeddings and report the result for the best performing embedding; Appendix~\ref{sec:appendix-training} lists the chosen embedding per corpus.

Across all collections, hETM achieves the lowest held-out perplexity at nearly every number of topics, with a clear margin over all baselines. On EUR-Lex the gap narrows: hETM, SawETM, and TSNTM form a tight cluster at every $K$, with hETM best at $K{=}10$ and $K{=}50$ (where it attains the overall lowest EUR-Lex perplexity), SawETM marginally lower at $K{=}20$ and $K{=}30$, with hETM and SawETM essentially tied at $K{=}100$; hETM has the lowest mean perplexity across $K$ on this corpus.

Among the hierarchical baselines, mlhLDA is the strongest model besides hETM on the U.S.\ Trade Statutes, Edgar, and 20 Newsgroups. It even edges hETM on Edgar at $K{=}10$, but degrades on EUR-Lex at larger $K$.
The flat embedded models are consistently weakest and stay roughly flat across $K$, with ETM(S) worst on the Trade Statutes and ETM(D) worst on Edgar. ETM(D) and ETM(S) are trained under the same protocol as hETM; their near-flatness across $K$ reflects a flat likelihood saturating on sectioned text, not undertraining. The remaining hierarchical baselines TSNTM and SawETM fall between hETM and mlhLDA, though TSNTM is unstable at large $K$, spiking on Edgar and 20 Newsgroups at $K{=}100$. Overall, hETM’s perplexity tends to improve or hold as $K$ grows, indicating that coupling the section and document topic distributions provides a better fit to these sectioned collections than either flat or topic hierarchy models. On three of the four corpora, mlhLDA $\succ$ flat ETM(D)/ETM(S) further supports that document--section coupling by itself improves fit, and hETM $\succ$ mlhLDA isolates the gain contributed by the embedding-based likelihood on top of that coupling. 

Beyond how well a model fits held-out text, topic models are also evaluated on the quality of the topics they produce. Two metrics are standard: topic coherence, computed as NPMI over the top 10 words from document level co-occurrence \citep{lau2014machine}, and topic diversity, the proportion of unique words among the top words of all topics \citep{dieng2020etm}. Following \citet{dieng2020etm}, we combine them into a single measure, topic quality, defined as the product of coherence and diversity, where higher is better. 

\begin{figure*}[h]
  \centering
  \includegraphics[width=\textwidth]{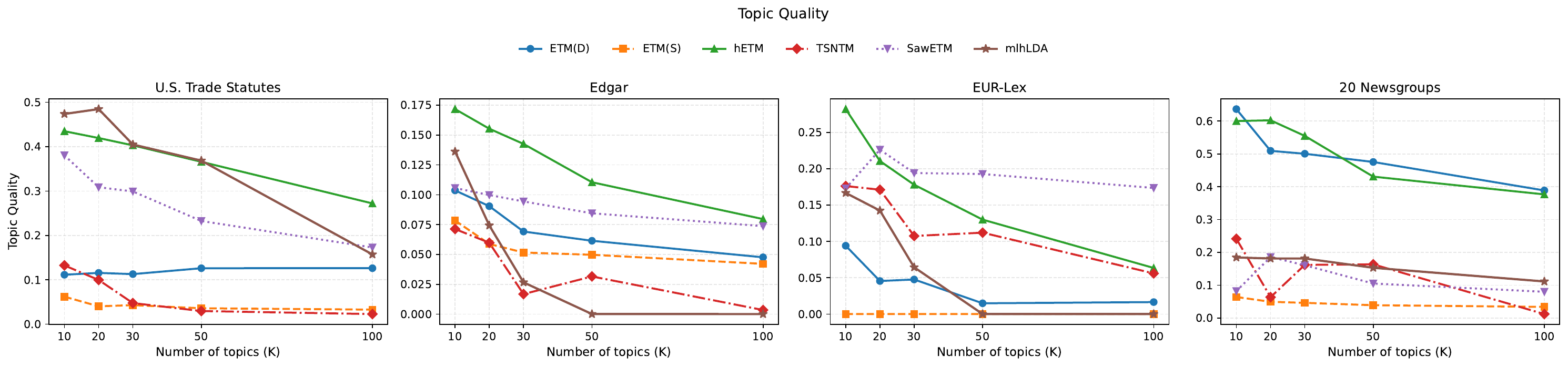}
  \caption{Topic quality across the four collections, for each model and number
  of topics $K$ (higher is better).}
  \label{fig:topic_quality}
\end{figure*}

Figure~\ref{fig:topic_quality} reports topic quality across the four collections. hETM produces the highest quality topics on Edgar at every number of topics. On the U.S.\ Trade Statutes, mlhLDA is the strongest model for $K{\le}50$ and hETM takes the lead at $K{=}100$ as mlhLDA’s quality collapses; across the Trade Statutes, mlhLDA and hETM are a clear pair above the remaining baselines. On EUR-Lex, SawETM is the strongest competitor and overtakes hETM for $K{\ge}20$, remaining high and roughly flat while hETM starts highest at $K{=}10$ but declines with $K$. On 20 Newsgroups, the two best models are hETM and the flat ETM(D), which trade the lead across $K$ (ETM(D) is highest at $K{=}10$, $50$, and $100$), while SawETM and TSNTM are much weaker there. Elsewhere, the flat and topic-hierarchy baselines are consistently the weakest, and ETM(S) even yields slightly negative quality on EUR-Lex. As is typical, topic quality falls as $K$ grows for most models. Table~\ref{tab:topic-examples} in Appendix~\ref{sec:example-topics} illustrates this qualitatively: hETM recovers thematically clean, distinctive topics on the U.S.\ Trade Statutes, while the baselines tend to repeat procedural boilerplate or fuse separate themes.

\section{Extrinsic Evaluation}
\label{sec:extrinsic}

A key advantage of topic models is that they represent every document in a shared, low-dimensional topic space, allowing documents with different vocabularies to be compared and used as features in downstream tasks. Beyond the intrinsic evaluations above, we therefore evaluate hETM extrinsically on a document clustering task, using the inferred document-topic representations as features and comparing hETM against the baseline models ETM(D), ETM(S), SawETM, TSNTM, and mlhLDA with k-means clustering.
For ETM(S), which treats each section as a separate document, we aggregate the per-section topic distributions by mean pooling to obtain a document representation. We measure cluster quality with the Adjusted Rand Index \citep[ARI;][]{hubert1985comparing}, a chance-corrected measure of agreement between the predicted clusters and the ground-truth labels that is robust to the label imbalance in the U.S.\ Trade Statutes and EUR-Lex corpora, where unadjusted accuracy can over-reward predictions that collapse onto the majority class; clustering accuracy and Normalized Mutual Information \citep[NMI;][]{strehl2002cluster} are reported in Appendix~\ref{sec:extrinsic-additional}. We run the task on three of the four collections, excluding 20 Newsgroups: there the hierarchy is implicit, with each newsgroup treated as a single document, leaving only 20 documents to cluster.

The labels differ by collection: for the U.S.\ Trade Statutes we evaluate clustering on the subset of laws that we assign to one of four trade-specific categories; for Edgar we use the Standard Industrial Classification (SIC) code, which places each company in one of eight sectors; and for EUR-Lex we use the EuroVoc top-level classification to assign each law to one of four trade categories. Table~\ref{tab:clustering} reports the best ARI per model and the $K$ at which it is attained.

\begin{table}[t]
\centering
\small
\setlength{\tabcolsep}{4pt}
\begin{tabular}{l ccc}
\toprule
Model          & U.S. Trade Statutes       & Edgar                     & EUR-Lex \\
\midrule
hETM           & \textbf{0.190}$_{20}$    & \textbf{0.334}$_{10}$    & 0.174$_{20}$          \\
SawETM         & 0.141$_{50}$             & 0.163$_{10}$             & \textbf{0.250}$_{10}$ \\
TSNTM          & 0.158$_{10}$             & 0.332$_{100}$            & 0.162$_{50}$          \\
mlhLDA         & 0.037$_{10}$             & 0.326$_{50}$             & 0.152$_{100}$         \\
ETM(D)         & 0.060$_{10}$             & 0.289$_{100}$            & 0.103$_{100}$         \\
ETM(S)         & 0.072$_{50}$             & 0.242$_{100}$            & 0.145$_{100}$         \\
\bottomrule
\end{tabular}
\caption{Document clustering ARI across the three collections (best per collection in bold). For each model, we select the number of topics $K$ that maximizes ARI; subscripts give that configuration. hETM is best on the U.S.\ Trade Statutes and Edgar; SawETM is best on EUR-Lex.}
\label{tab:clustering}
\end{table}

hETM achieves the highest ARI on the U.S.\ Trade Statutes and Edgar datasets, attaining these results at relatively low topic counts ($K{=}20$ and $K{=}10$). In contrast, baseline models reach peak performance only at substantially higher $K$ values, such as $K{=}100$ for both TSNTM and ETM(D) on Edgar. On EUR-Lex, hETM remains competitive but does not outperform all baselines; SawETM achieves the highest ARI, with hETM second. This is consistent with the intrinsic evaluation, where SawETM also performs best on this corpus. Among the hierarchical baselines, mlhLDA tracks its intrinsic ranking across corpora, and the hETM--mlhLDA gap again isolates the gain from the embedding-based likelihood. No baseline consistently ranks second: ETM(D) is weakest on the Trade Statutes, while TSNTM and SawETM alternate elsewhere. Appendix~\ref{sec:extrinsic-additional} reports accuracy and NMI under the same per-metric best-$K$ selection; all three metrics give the same ranking per corpus, so the ranking is not an artifact of a single metric or of label imbalance.

\section{Conclusion}
\label{sec:conclusion}
We introduced hETM, a generative hierarchical embedded topic model for sectioned documents that draws each section’s topic proportions as a Gaussian perturbation of its document’s in logit space. The coupling keeps the model reparameterizable and yields a closed-form level-coupling term fit by amortized variational inference. Across legal, financial, and online forum corpora, hETM attains the lowest held-out perplexity at every $K$ on the U.S.\ Trade Statutes and 20 Newsgroups, wins four of five $K$ on Edgar (ceding only $K{=}10$ to mlhLDA), and achieves the single lowest EUR-Lex perplexity point ($K{=}50$). It also attains the best topic quality on Edgar at every $K$ and the best clustering on the U.S.\ Trade Statutes and Edgar, with mlhLDA the strongest topic-quality competitor at small $K$ on the Trade Statutes and SawETM the strongest competitor on EUR-Lex. These results support the benefit of coupling section and document topics rather than modeling either level alone.

\section*{Limitations}
\label{sec:limitations}

hETM assumes documents are partitioned into sections; when the partition is implicit, we construct it heuristically. Like other topic models, hETM relies on bag of words representation and ignores word order within a section. The hierarchical logistic normal prior approximates rather than exactly mirrors the nested Dirichlet of mlhLDA, since the correspondence holds only in the logit space. And like other amortized variational models, hETM is subject to the usual amortization gap: a shared inference network cannot reach the per document posterior of mean field variational inference.

The variational posterior factorizes across sections (Section~\ref{sec:inference}), so each section posterior $q_{\nu_s}(\delta^s_{i,j}\mid x_{i,j})$ reads only its own section's word counts. The coupling between the document and section levels enters the ELBO only through the KL term that pulls the posterior means together. A variational family of the form $q(\delta^s_{i,j}\mid x_{i,j}, \delta^d_i)$, which conditions each section posterior on the document logit, would more faithfully mirror the generative process and is a natural direction for future work.

Although a document is defined as an ordered list of sections, the model treats sections as conditionally independent given the document representation. Properly exploiting section order would require a sequential prior over section logits and a correspondingly structured inference network. We leave this to future work.

Finally, our corpora are English and our hierarchies are two levels deep, so extensions to multilingual corpora and to hierarchies with more than two levels remain unevaluated.

\bibliography{hetm}

\clearpage
\appendix
\section{Example Topics}
\label{sec:example-topics}

\begin{strip}
\centering
\small
\renewcommand{\arraystretch}{1.5}
\setlength{\tabcolsep}{4pt}

\begin{tabular}{@{}c l c c l@{}}

toprule
Theme & Model & C & D & Top words \\
\midrule
\multirow{6}{*}{\rotatebox[origin=c]{90}{\parbox{2.6cm}{\centering Agriculture \& commodities}}}
 & \textbf{hETM} & \textbf{0.49} & \textbf{1.00} & farm agricultural crop food allotment producer marketing wheat \\
 & SawETM & 0.33 & 0.80 & good acreage producer farm wheat marketing allotment usmca \\
 & TSNTM & 0.23 & 0.00 & trade agricultural export person country bank international provisions \\
 & ETM(D) & 0.24 & 0.00 & house senate approved articles june considered passed representatives \\
 & ETM(S) & 0.13 & 0.50 & trade international zinc cotton export extension copper air \\
 & mlhLDA$^\dagger$ & 0.33 & 1.00 & control fire adoption anti licensed horses medication doping \\
\midrule
\multirow{6}{*}{\rotatebox[origin=c]{90}{\parbox{2.6cm}{\centering Health \& family}}}
 & \textbf{hETM} & \textbf{0.34} & \textbf{1.00} & care medical drug child children family treatment prevention \\
 & SawETM & 0.20 & 0.20 & contract care pay acquisition projects operations medical property \\
 & TSNTM & 0.61 & 1.00 & alumina fungible molds spp isoxaflutole chicory tetrafluoroethylene naphthalenyl \\
 & ETM(D) & 0.24 & 0.00 & house senate approved articles june considered passed representatives \\
 & ETM(S) & 0.15 & 0.00 & representatives amendments reports ways treatment means provisions administrator \\
 & mlhLDA$^\dagger$ & 0.49 & 0.90 & diseases soil severe extreme infections pathogens infection unicef \\
\midrule
\multirow{6}{*}{\rotatebox[origin=c]{90}{\parbox{2.6cm}{\centering Banking \& finance}}}
 & \textbf{hETM} & \textbf{0.30} & \textbf{0.90} & business bank small corporation financial loan investment credit \\
 & SawETM & 0.22 & 0.30 & export bank corporation credit company international policy trade \\
 & TSNTM & 0.23 & 0.00 & trade agricultural export person country bank international provisions \\
 & ETM(D) & 0.24 & 0.00 & house senate approved june articles considered representatives passed \\
 & ETM(S) & 0.18 & 0.20 & trade america bank international tariff free credit total \\
 & mlhLDA$^\dagger$ & 0.33 & 0.50 & tralkoxydim effect barium resolution joint glass covered modulus \\
\bottomrule

\end{tabular}

\captionof{table}{Example topics from each model on the U.S.\ Trade Statutes corpus, grouped by theme. For each topic, we report coherence (C) and topic diversity (D), the fraction of a topic's top ten words that no other topic in the same model uses, together with its top words. hETM recovers thematically clean, distinctive topics, whereas the baselines either contaminate the theme with procedural boilerplate (ETM(D)), fragment it (ETM(S)), or fuse it with unrelated provisions (SawETM, TSNTM). The apparent high coherence of TSNTM in the health row comes from a degenerate topic of rare chemical tokens, not a genuine theme. $^\dagger$At $K=50$ mlhLDA has no topic matching these themes; we show its nearest topic, whose coherence reflects a narrow, off-theme product cluster (anti-doping and licensing, plant pathogens, and a chemical description, respectively) rather than the theme itself.}

\label{tab:topic-examples}
\end{strip}
\clearpage
Table~\ref{tab:topic-examples} shows example topics recovered by each model on the U.S.\ Trade Statutes corpus, grouped by theme, with coherence (C) and diversity (D) as defined in Section~\ref{sec:intrinsic}. hETM recovers thematically clean, distinctive topics, whereas the baselines either contaminate the theme with procedural boilerplate (ETM(D)), fragment or fuse it (TSNTM, SawETM), or produce low-coherence mixtures (ETM(S)). mlhLDA at $K=50$ recovers none of these themes, instead splitting the vocabulary into narrow product clusters (for example anti-doping and licensing, plant pathogens, and chemical descriptions), and only produces a clean agriculture topic at $K=100$ --- an illustration of how its free categorical topic-word distributions fragment the topic space.
\clearpage

\section{Section Variance Ablation}
\label{sec:sigma-ablation}

The section variance $\sigma_s^2$ in hETM controls how tightly each section's topic distribution is pulled toward its document's. Two limiting cases bracket its role: $\sigma_s^2 \to 0$ makes every section share its document's mixture, reducing the reconstruction term to that of ETM(D), while $\sigma_s^2 \to \infty$ eliminates this coupling entirely, so sections are fit independently, closer to ETM(S). To check that the coupling does real work in between, we sweep $\sigma_s^2 \in \{0.1,\,1,\,10\}$ (small, default, large) and retrain hETM at $K \in \{10,20,30,50,100\}$ on two corpora chosen for their contrasting section topic structure: the U.S.\ Trade Statutes, where sections of a single bill often address different subtopics, and EUR-Lex, where sections of a single regulation tend to stay on a single theme.

\begin{figure}[h]
  \centering
  \includegraphics[width=\columnwidth]{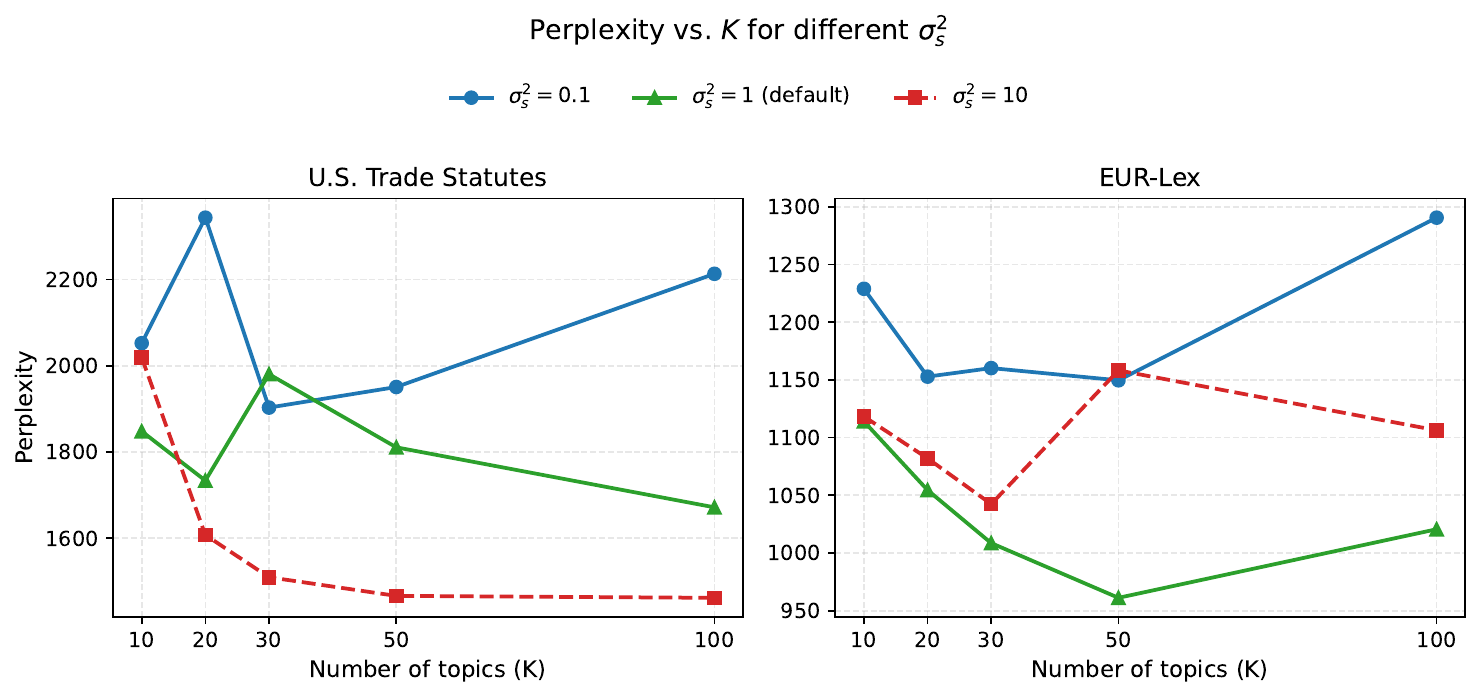}
  \caption{Held-out perplexity of hETM as a function of the number of topics $K$ for $\sigma_s^2 \in \{0.1,\,1,\,10\}$ on the U.S.\ Trade Statutes and EUR-Lex. The default $\sigma_s^2 = 1$ used throughout the paper is in green.}
  \label{fig:sigma-ablation-perplexity}
\end{figure}

\begin{figure}[h]
  \centering
  \includegraphics[width=\columnwidth]{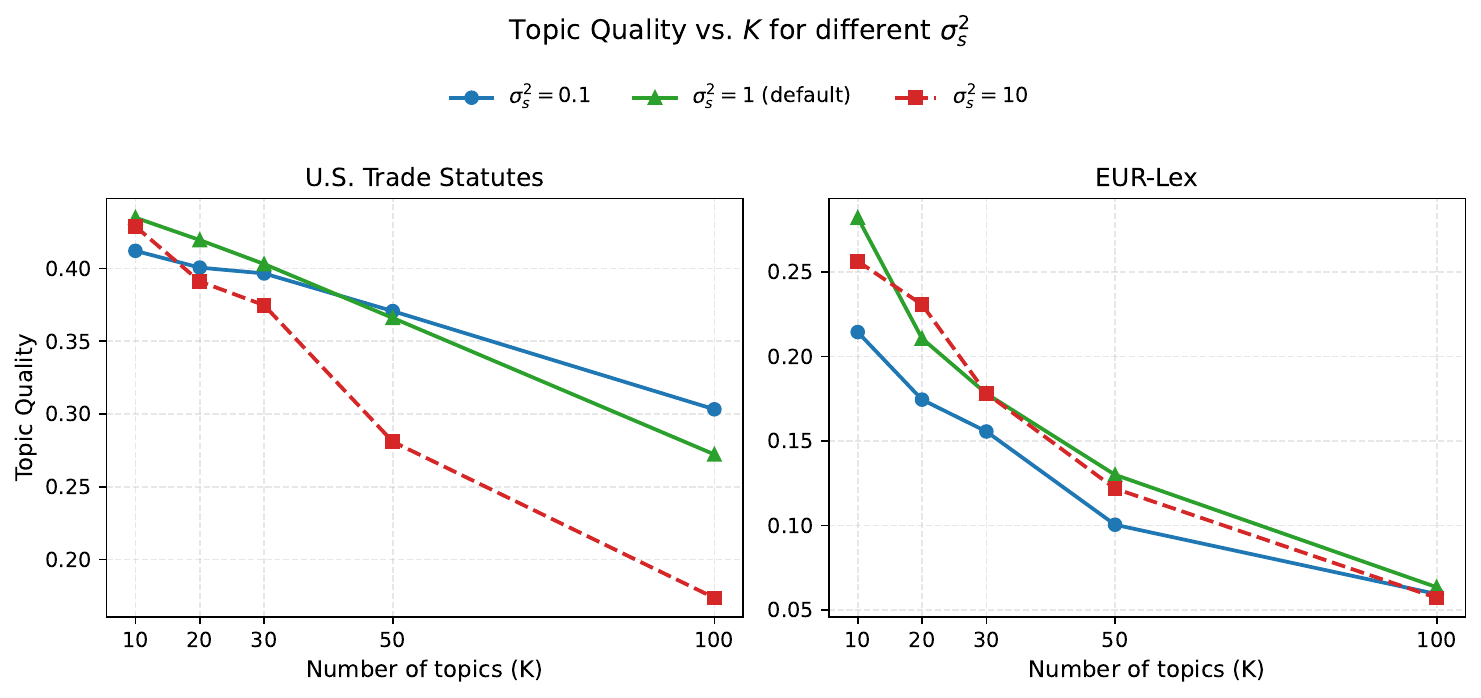}
  \caption{Topic quality (coherence multiplied by diversity) for the same ablation. Higher is better.}
  \label{fig:sigma-ablation-topic-quality}
\end{figure}

Figures~\ref{fig:sigma-ablation-perplexity} and \ref{fig:sigma-ablation-topic-quality} show the sweep. The two corpora behave differently, and the pattern lines up with how much section topic variation each contains. On EUR-Lex the three curves stay close together on both metrics, and the default $\sigma_s^2 = 1$ gives the lowest perplexity at every $K$ and the highest topic quality at most $K$. On the U.S.\ Trade Statutes the three curves fan out, and the sweep exposes a trade off: the loose extreme $\sigma_s^2 = 10$ gives the lowest perplexity at $K \geq 20$ while the tight extreme $\sigma_s^2 = 0.1$ gives the highest topic quality at $K \geq 20$; the default $\sigma_s^2 = 1$ sits between the two.

This is the regime hETM is designed for: because sections of a single bill address substantially different subtopics, letting each section drift ($\sigma_s^2 = 10$) throws away the shared document theme and collapses topic quality at large $K$, while tightening too far ($\sigma_s^2 = 0.1$) forces sections into the document's average mixture and hurts perplexity. The trade off is itself informative. On corpora where sections genuinely cover different subtopics, the choice of $\sigma_s^2$ trades section level fit against collection level topic coherence, and no single value dominates. On corpora where section level variation is small, this trade off collapses and the default suffices. In the main paper experiments we fix $\sigma_s^2 = 1$ as a balanced default. Algorithm~\ref{alg:hetm} notes that the variances can optionally be learned, but we did not do so in Sections~\ref{sec:intrinsic} and \ref{sec:extrinsic}, so this ablation isolates a single hyperparameter.

We did not rerun the full sweep on Edgar or 20 Newsgroups. The pattern on the two corpora shown here, in which sensitivity to $\sigma_s^2$ scales with the amount of section topic variation in a corpus, predicts that these two sit between the extremes shown; we leave empirical verification to future work.
\clearpage
\section{Clustering Accuracy and NMI}
\label{sec:extrinsic-additional}

For completeness, and to confirm that the ARI ranking in Table~\ref{tab:clustering} is not an artifact of a single metric, Table~\ref{tab:clustering-accuracy} and Table~\ref{tab:clustering-nmi} report clustering accuracy and Normalized Mutual Information (NMI) \citep{strehl2002cluster} for the same models and corpora. For each table and each model we select the number of topics $K$ that maximizes the reported metric and give it as a subscript, as in the main-body ARI table.

\begin{table}[h]
\centering
\small
\setlength{\tabcolsep}{4pt}
\begin{tabular}{l ccc}
\toprule
Model          & U.S. Trade Statutes       & Edgar                     & EUR-Lex \\
\midrule
hETM           & \textbf{0.582}$_{20}$    & \textbf{0.571}$_{10}$    & 0.512$_{20}$          \\
SawETM         & 0.546$_{50}$             & 0.446$_{10}$             & \textbf{0.528}$_{10}$ \\
TSNTM          & 0.500$_{10}$             & 0.564$_{100}$            & 0.472$_{50}$          \\
mlhLDA         & 0.409$_{50}$             & 0.524$_{50}$             & 0.521$_{20}$          \\
ETM(D)         & 0.473$_{20}$             & 0.557$_{100}$            & 0.474$_{100}$         \\
ETM(S)         & 0.470$_{30}$             & 0.512$_{100}$            & 0.449$_{100}$         \\
\bottomrule
\end{tabular}
\caption{Document clustering accuracy across the three collections (best per collection in bold). For each model we select the number of topics $K$ that maximizes accuracy; subscripts give that configuration.}
\label{tab:clustering-accuracy}
\end{table}

\begin{table}[h]
\centering
\small
\setlength{\tabcolsep}{4pt}
\begin{tabular}{l ccc}
\toprule
Model          & U.S. Trade Statutes       & Edgar                     & EUR-Lex \\
\midrule
hETM           & \textbf{0.236}$_{20}$    & \textbf{0.402}$_{100}$   & 0.190$_{20}$          \\
SawETM         & 0.222$_{10}$             & 0.362$_{10}$             & \textbf{0.258}$_{10}$ \\
TSNTM          & 0.186$_{50}$             & 0.387$_{100}$            & 0.158$_{30}$          \\
mlhLDA         & 0.206$_{20}$             & 0.320$_{10}$             & 0.234$_{50}$          \\
ETM(D)         & 0.219$_{30}$             & 0.324$_{100}$            & 0.185$_{100}$         \\
ETM(S)         & 0.181$_{10}$             & 0.245$_{100}$            & 0.104$_{30}$          \\
\bottomrule
\end{tabular}
\caption{Document clustering Normalized Mutual Information (NMI) across the three collections (best per collection in bold). For each model we select the number of topics $K$ that maximizes NMI; subscripts give that configuration.}
\label{tab:clustering-nmi}
\end{table}
\clearpage

\section{Training Configuration and Reproducibility}
\label{sec:appendix-training}

In this appendix we provide the training configuration used for all hETM and ETM results in the paper, along with the pretrained-embedding selection for hETM. ETM is the standard ETM implementation of \citet{dieng2020etm} with its default settings. hETM uses two separate variational encoders, one at the document level and one at the section level, each with two fully connected hidden layers of 800 units and ReLU activations, taking normalized bag-of-words inputs and parameterizing diagonal Gaussian posteriors. The document and section encoders share the global ETM topic-word decoder, with topic embeddings of the same dimension as the word embeddings (768 by default, 512 when LegalBERT-small is used). All models are trained with Adam, learning rate 0.005, weight decay $1.2\times 10^{-6}$, and no dropout, with early stopping. hETM uses minibatches of 16 parent documents and processes sections in chunks of up to 64 (USTS, Edgar) or 128 (EUR-Lex) per document; ETM uses minibatches of 32 documents on USTS and Edgar and 64 on EUR-Lex. We use a 90/10 train/test split taken at the parent-document level with preparation seed 42; training uses seed 2019. 

Pretrained word embeddings (LegalBERT, InCaseLawBERT, FinBERT, ModernBERT; dimensions 768 or 512) are kept frozen during optimization. The learned configuration initializes embeddings randomly and trains them jointly with the topic model. Pretrained word vectors are obtained following \citet{bommasani2020interpreting}: for each vocabulary word, we collect its contextual representation at every occurrence in the corpus and average these vectors to a single static embedding per type. 

Section~\ref{sec:word_embeddings} introduces six embedding configurations for hETM (LegalBERT-base, LegalBERT-small, InCaseLawBERT, FinBERT, ModernBERT, and learned). We train hETM once per configuration and the main text reports the configuration with the lowest held-out perplexity. Table~\ref{tab:best-embedding} lists the winning configuration on each corpus.

\begin{table}[h]
\centering
\small
\setlength{\tabcolsep}{4pt}
\begin{tabular}{l l}
\toprule
Corpus              & Best embedding for hETM \\
\midrule
U.S. Trade Statutes & Learned \\
Edgar               & Learned \\
EUR-Lex             & LegalBERT-base \\
20 Newsgroups       & Learned \\
\bottomrule
\end{tabular}
\caption{Pretrained (or learned) word embedding configuration that achieved the lowest held-out perplexity for hETM on each corpus. Six configurations are compared; see Section~\ref{sec:word_embeddings} for the configuration details.}
\label{tab:best-embedding}
\end{table}

On three of the four corpora the learned embedding wins; on EUR-Lex LegalBERT-base wins. We attribute this to corpus size and vocabulary coverage: EUR-Lex is our largest corpus and its legal vocabulary overlaps substantially with LegalBERT's pretraining, so the pretrained embedding wins over a learned one even after mean-pooling its subword pieces. On the smaller corpora, and on 20 Newsgroups whose informal vocabulary matches none of our pretrained models, a learned embedding has more freedom to adapt to the actual vocabulary and outperforms every pretrained option we tested.
\clearpage
\section{Parameter Counts and Training Time}
\label{sec:appendix-cost}

hETM uses a second variational encoder for the section level, which roughly doubles the encoder parameter count relative to ETM. The topic word decoder is shared across the two levels, so the full hETM parameter count stays below twice that of ETM at the same vocabulary and topic count. Training time per epoch grows with the mean number of sections per document: hETM is under two times slower than ETM on Edgar (mean 3.4 sections per document) and about five times slower on the U.S.\ Trade Statutes (mean 69 sections per document). At inference time hETM applies both encoders, so inference latency per document is roughly twice ETM's. All runs reported in the main text, across all four corpora and all topic counts, fit within 24 hours of wall clock time on a single Nvidia RTX 2080TI GPU.

\end{document}